\documentclass{llncs}
\usepackage{makeidx}  % allows for indexgeneration

\usepackage{epsfig}
\usepackage{subfigure}
\usepackage{ upgreek }
\usepackage{graphicx}
\usepackage{amsmath}
\usepackage{amssymb}
\usepackage{textcomp}
\usepackage{hyperref}
\usepackage{color}

\begin{document}
% 
%\frontmatter          % for the preliminaries
%
\pagestyle{headings}  % switches on printing of running heads
%\addtocmark{Hamiltonian Mechanics} % additional mark in the TOC
%
%
\mainmatter              % start of the contributions
\title{V-FCNN: Volumetric Fully Convolution Neural Network For Automatic Atrial Segmentation}

%\iffalse
\author{Nicol\'o Savioli$^{1}$,Giovanni Montana$^{1,2}$,Pablo Lamata$^{1}$}%\fi 

\institute{Department of Biomedical Engineering, King’s College London, SE1 7EH, UK \\ \email{\{nicolo.l.savioli,pablo.lamata,giovanni.montana\}@kcl.ac.uk}\and
WMG, University of Warwick, Coventry, CV4 71AL \\
\email{g.montana@warwick.ac.uk}}

\maketitle

\begin{abstract}

Atrial Fibrillation (AF) is a common electro-physiological cardiac disorder that causes changes in the anatomy of the atria. A better characterization of these changes is desirable for the definition of clinical biomarkers, furthermore, thus there is a need for its fully automatic segmentation from clinical images.  
In this work, we present an architecture based on 3D-convolution kernels, a Volumetric Fully Convolution Neural Network (V-FCNN), 
able to segment the entire volume in a one-shot, and consequently integrate the implicit spatial redundancy present in high-resolution images. A loss function based on the mixture of both Mean Square Error (MSE) and Dice Loss (DL) is used, in an attempt to combine the ability to capture the bulk shape as well as the reduction of local errors products by over-segmentation. Results demonstrate a reasonable performance in the middle region of the atria along with the impact of the challenges of capturing the variability of the pulmonary veins or the identification of the valve plane that separates the atria to the ventricle. A final dice of $92.5\%$ in $54$ patients ($4752$ atria test slices in total) is shown.

\end{abstract}

\section{Introduction}

Atrial Fibrillation (AF) is a common electro-physiological cardiac disorder with a large prevalence worldwide \cite{Prystowsky} that causes changes in the anatomy of the atria. A better characterization of these changes as a consequence of the substrate that causes and sustains fibrillation is desirable for the definition of clinical biomarkers. These biomarkers can be directly started from the image (i.e. the shape of the atria \cite{Varela17}, or the fibrotic burden from late gadolinium enhanced (LGE) magnetic resonance imaging (MRI) \cite{Kim}) or from mechanistic simulations of the function (i.e. computation of the risk of arrhythmia perpetuation \cite{Boyle16}). 

There is thus a need for a fully automatic segmentation of the atria from clinical images, especially in LGE studies. The current state of the art is based in tedious along with error-prone manual procedures, and the main difficulty is the lack of contrast from atrial tissue including the surrounding background. Fully automated solutions are desirable to speed up the process and remove inter- and intra-observer variability. In this direction, a combination of multi-atlas registration within 3D level-set has been proposed, reporting a reasonable performance in the main atrial body and pulmonary vein regions \cite{Tao}. The large computational burden of this multi-atlas approach can be alleviated by the use of convolutional neural networks (CNNs), as it has been illustrated in \cite{Aliasghar} for the analysis of 2D MRI slices. 

The first idea followed in this work is the use of 3D CNNs as the effective solution for the segmentation of the 3D LGE datasets. The starting point is the v-net \cite{Fausto}, which takes into consideration the spatial redundancy naturally present on the entire volumetric stack with 3D-kernels, showing good benefits in different cardiac segmentation problems \cite{Isensee},\cite{Hinrich}. This architecture is modified to reduce the memory burden and speed-up its training. The last idea explored in this work is the choice of a sensible loss function, the joint combination of both the mean squared error (MSE) and Dice Loss, which has been reported to be beneficial as the MSE minimizes global image details while the Dice Loss reduces local over-segmentation errors \cite{Fausto}.

\begin{figure*}[ht]
\centering
 \includegraphics[width=5in]{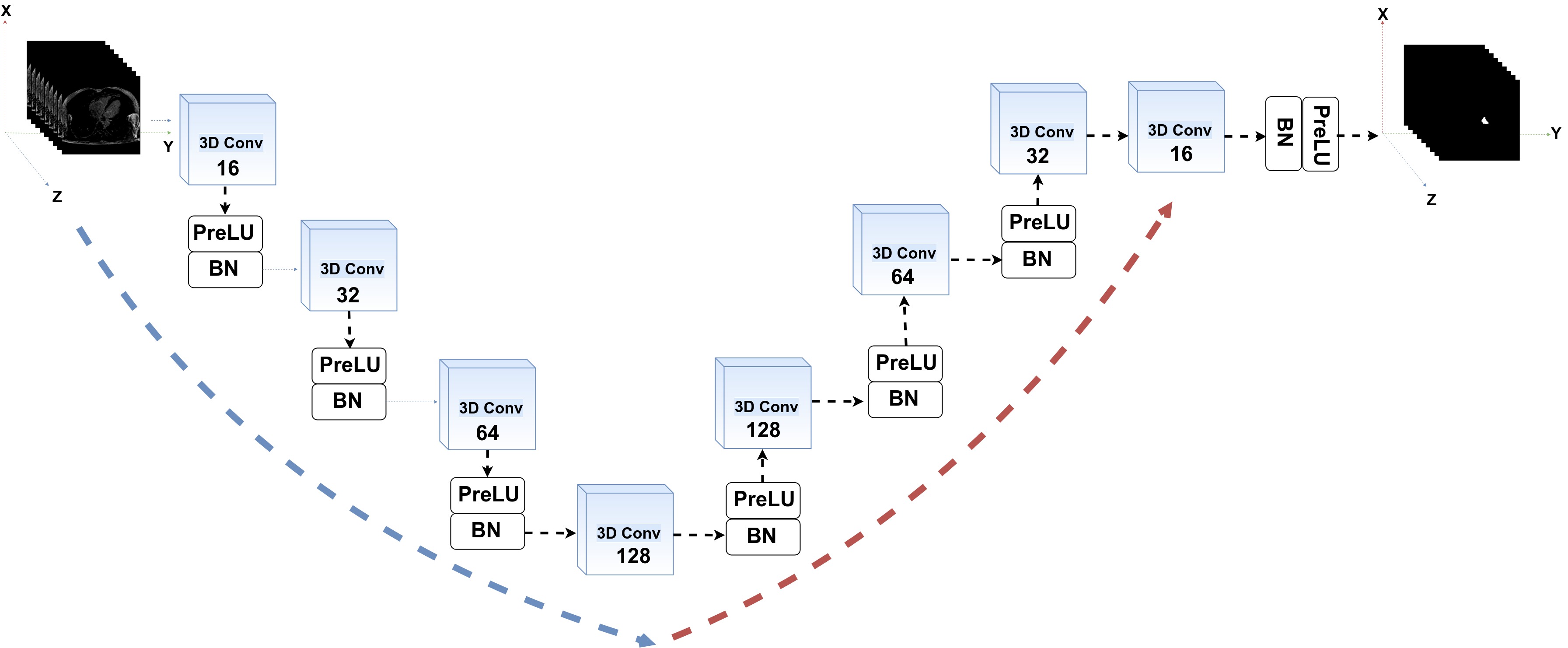}
   \caption{V-FCNN architecture. Input is the (XYZ) 3D MRI volume of size ($127\times127\times 88$), also passed through the down-sampling path (blue arrow),
represented by a 3D kernels Convolution Neural Network (CNN) able to progressively reduce the input volume slices.
Then, the hidden features, at the end of it, are restored within 3D up-sampling kernels (red-arrow), ending in an output being a 3D mask of size ($127\times127\times 88$). Both down-sampling and up-sampling paths consist of four 3D-convolutions blocks (blue boxes) followed by PreLU plus 3D-Batch Normalisation (BN). The number of feature maps for each convolution layers are $16,32,64,128$ both in down and up-sampling.}
\label{fig:fig1}
\end{figure*}

\section{Atrial Datasets}

A population of 100 3D GE-MRIs, data and masks, were made available through the [\href{http://atriaseg2018.cardiacatlas.org/}{2018 Atrial Segmentation Challenge}] and used in this work. Images had a acquisition resolution of 0.625 x 0.625 x 0.625 $mm^3$. 

\section{Method}

A volumetric fully convolutional network, V-FCNN, is designed with two main paths (see Fig.1): the volumetric down-sampling path as well as the volumetric up-sampling path. The volumetric down-sampling path has four 3-D convolutions blocks, each following by PreLU along with 3-D Batch Normalisation (BN) layers. This path takes as input the entire (XYZ) volume and progressively reduces the size of each slice (XY) together with the number of stack spatial slices (Z). In this phase, the volume is compressed and presents both (XY) and (Z) reduction. 
In a complementary fashion, the volumetric up-sampling path restores the compressed volume to its initial size, with every 3-D up-sampling convolution blocks being followed by PreLU and 3-D Batch Normalisation (BN).
Each sampling paths of V-FCNN contains four blocks within 16,32,64,128 -3D kernels respectively (with the size fix to $3\times3$). 

Image down-sampling during a segmentation task 
presents the problem of feature map reduction, followed by a strong spatial information loss.
This problem has been addressed in v-net \cite{Fausto}, \cite{Isensee} by adding skips layers between down-sampling and up-sampling layers
in order to fuse low-level features within high level
features.

Our work explores an alternative approach to the skip path connections. In particular, we boost our model to capture fine details in two ways. 
First, the use of max-pooling operations is avoided in order to prevent the loss of spatial resolution (i.e if pools do not overlap well, pooling operation loses appreciable information where the objects are located in the image). The second idea is to use a loss layer that combines the $MSE_{loss}$ and $DICE_{loss}$ metrics as an attempt to recover details in the image. The rationale is that the $DICE_{loss}$ term searches for local details in the volume data, as a consequence, the $MSE_{loss}$ can be seen as a regularization that instead focuses on global features on the MRI volume. This bimodal loss also prevents the V-FCNN to fall in a local minimum; especially within small atrial regions. Specifically, the loss is defined as: 

\begin{equation}
\begin{split}
 Loss =  MSE_{loss} + \lambda\cdot DICE_{loss} =  \frac{1}{Z} \sum_{s=0}^{Z} (y[s]-\hat{y}[s])^{2} \\
  + \lambda\cdot  \frac{1}{Z} \sum_{s=0}^{Z} \frac{2\sum_{i}^{N} y[s]_{i}\hat{y}[s]_{i}}{\sum_{i}^{N} y[s]_{i} + \sum_{i}^{N} \hat{y}[s]_{i}},
\end{split}
\end{equation}

where $y[s]$ are the Ground Truth (GT) binary slices, as well as $\hat{y}[s]$ are the correspondent's prediction masks. $s$ is an index through the spatial slices in the Z (z-axis).
Then, the sums for $i$ in $DICE_{loss}$ runs over all N pixels of the prediction masks $\hat{y}[s]$ and the $y[s]$ GT masks. Finally, $\lambda$ controls the amount of $DICE_{loss}$ during the optimisation training process (set to $1e-3$). 

The size of each MRI slice is reduced to $127\times127$ pixels using down-sampling bi-cubic interpolation, allowing the network to be faster in training stage moreover enabling a solution without expensive GPU hardware (i.e. large images need more GPU global memory $\sim 32$ GB). An up-sampling bi-cubic interpolation is finally used for restoring the mask to the size of $640\times640$ pixels. 

 \begin{figure*}[ht]
\centering
 \includegraphics[width=4in]{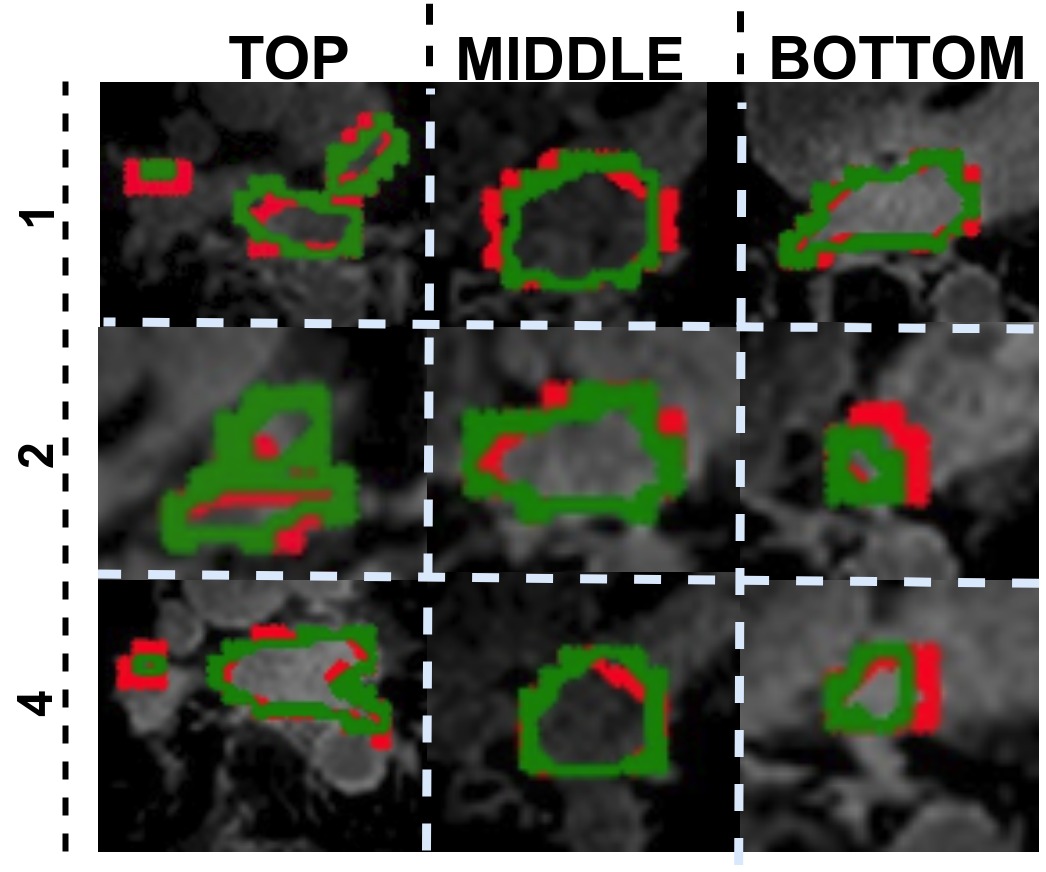}
   \caption{Visual comparison of the segmentations obtained from V-FCNN (green line) vs clinical ground truth (red line) in three different test patients (number 1, 2 or 4). The comparison is made at three different sections of the atrium: top, middle and bottom. Note how the V-FCNN is able to segment not only visually simpler slices (middle section) but also more complex cases (top and bottom sections).}
\label{fig:fig2}
\end{figure*}

\subsection{Implementation details}

For our experiments, we train the network with a number of epochs of 1000 up to convergence. The Stochastic Gradient Descent (SDG) was used with a  learning rate of $1e-4$, while the momentum and weight decay are $0.9$, $1e-5$ respectively. Furthermore, to increase the generalization of the network, data argumentation was used, finding particularly effective the random vertical and horizontal flip in close combination with plane translation. Input sequences were equalized in grayscale intensity with CLAHE (Contrast Limited Adaptive Histogram Equalization) \cite{Kaur}, and the noise was minimized with a combination of High-Pass Filters and Gaussian blurring filters.

\begin{figure*}[ht]
\centering
 \includegraphics[width=5in]{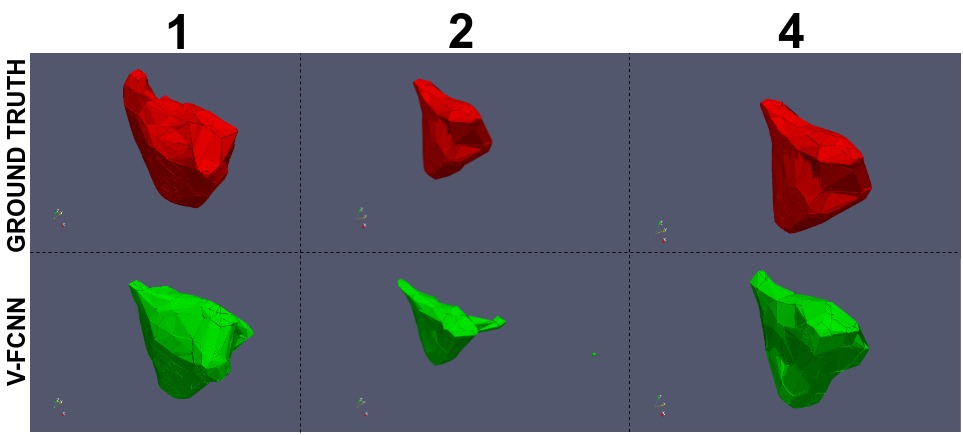}
   \caption{Visual comparison between ground truth  (red) compared with those obtained by proposed V-FCNN (green). Note that, the mesh coarse resolution is related to the low number of triangles used.}
\label{fig:result_2}
\end{figure*}

\section{Experiments}

The experimentation phase is divided into two phases. In the first phase (preparation phase) 5 patients are used for validation ($440$ slices in total), and the rest (90) is used for training. In the second phase (competition), our algorithm was evaluated in the test-set with 54 patients ($4752$ atria slices in total).

Segmentation accuracy was measured with the Dice Metric (DM) \cite{Avendi}, which was subdivided into 3 regions of the atria: top (including the pulmonary veins), middle and bottom (including the valve plane that divides the atria and ventricle). Besides, DM and the surface Hausdorff Distance (HD) \cite{Cignoni} are computed for the entire atrial anatomy.

Particularly, given a point $p$ and a surface $S$, the distance $\upvarepsilon(p.S)$ is defined as: 

\begin{equation}
\upvarepsilon(p,S)= min_{p^{'} \in S} d(p,p^{'})
\end{equation}

whereas the $d(\cdot)$ is the Euclidean distance between two points in a Euclidean space.
Then, the HD between mesh surfaces $S_{GroundTruth}$ and $S_{vfcnn}$ is given as: 

\begin{equation}
HD(S_{GroundTruth},S_{vfcnn}) = max_{p \in S_{GroundTruth}} \upvarepsilon(p,S_{vfcnn})
\end{equation}

Proposed V-FCNN achieved in the first phase an average DM of $69.6 \pm 16.1$, $82.1 \pm 1.4$ and $78.0 \pm 6.0$ in the top, middle and bottom regions, respectively (see Table 1 and an illustrative example in Fig. 2). The HD ranged from 0.31 to 0.86 mm. The average DM and HD were $76.58(7.87)$ and $0.59$ respectively. The DM in the competition phase was $92.5\%$.

Visual inspection of the contours reveals how a loss of accuracy is seen on the first top slices for two of the cases, creating an artificial flattening of the shape (see patients 1 and 2 of Fig. 3), and on the top of the atria with the more variable anatomy of the pulmonary veins. 

\begin{table}[]
\centering
\begin{tabular}{|l|l|l|l|l|}
\hline
                        & \textit{\textbf{TOP}}  & \textit{\textbf{MID}} &  \textit{\textbf{BOTTOM}}   & \textit{\textbf{3D-MESH}} \\ \hline
\textbf{Patient number} & \textbf{DM (\%)}       & \textbf{DM (\%)}      & \textbf{DM (\%)}       & \textbf{HD}          \\ \hline
1                       &  77.74 (8.46)           & 84.62 (2.40)          &   \textbf{76.99 (10.89)}   & 0.86                      \\ \hline
2                       &  72.26 (9.50)           & 81.25 (1.57)          &   54.18 (33.70)            & 0.66                      \\ \hline
3                       &74.10 (8.92)             & 77.75 (1.36)          &   68.27 (14.75)           & \textbf{0.31}             \\ \hline
4                       & 81.60 (0.74)            & 81.54 (1.20)          &   72.32 (14.32)        & 0.55                      \\ \hline
5                       &\textbf{84.48 (2.58)}    & \textbf{85.36 (0.66)} &   76.33 (7.05)  & 0.58                      \\ \hline
\end{tabular}
\newline
\caption{Automatic segmentation results (Dice Metric and Hausdorff Distance) for all five test patients. Results report the mean and standard deviation, and are divided into three different atrium sections: top, middle and bottom.}
\end{table}

\begin{figure*}[ht]
\centering
 \includegraphics[width=4in]{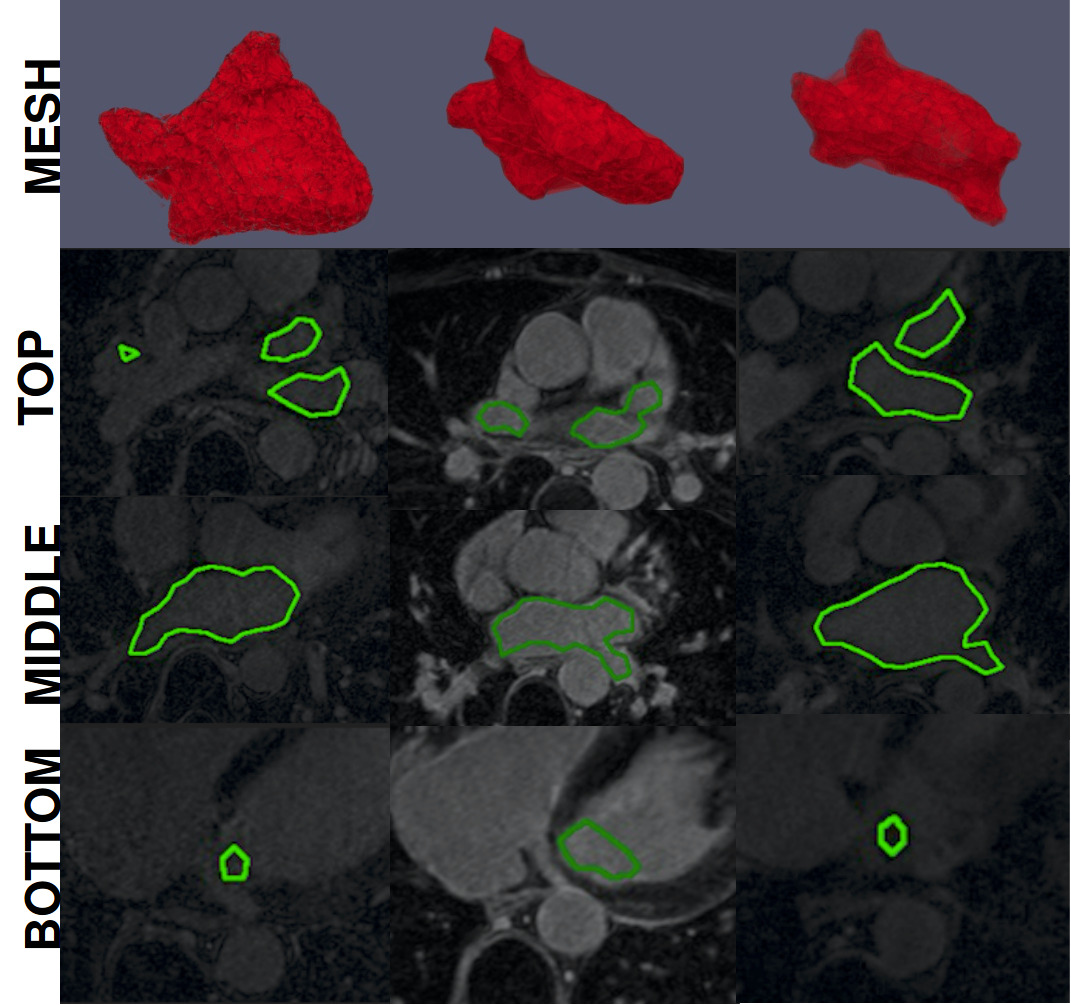}
   \caption{Exemplary result in 3 cases of the competition cases, illustrated with the 3D reconstruction (red, top) and 3 different sections of atrium segmentation: top, middle, and bottom.}
\label{fig:final_1}
\end{figure*}

\section{Discussion and Conclusions}

The exploitation of spatial coherence across different volumetric slices is an important resource for improving the accuracy of fully automated segmentation. Proposed V-FCNN achieves a good segmentation performance, mostly in the middle atrial section. Limitations occur in the top and bottom sections, caused by the presence of the pulmonary veins and the difficulty to identify where the atria and ventricle split. In the preparation phase the V-FCNN  achieved a DM between $82.05 (3.43)$ (top case) and $69.23 (14.92)$ (worst case), along with a DM of $92.5$ at the competition phase.

A cardiac imaging study has a huge level of correlation in space, as well as many segmentation algorithms have exploited the spatial coherence within different deep-learning techniques \cite{Fausto}, \cite{Isensee}, \cite{Hinrich}, \cite{Poudel}, \cite{Jianxu}. Some of them have used 3D convolutional kernels \cite{Fausto}, \cite{Isensee}, \cite{Hinrich} that have the advantage of directly capturing the spatial information in each convolutional layer without adding extra parameters (i.e. the addition of a recurring network). This is the main reason that has driven us to use them in our optimized V-FCNN. 

The limitation of the 3D convolutional kernels is the large memory burden, and this is the reason why the original detail of the images was reduced (from 640 to 127 pixels). 
However, this image reduction is required due to the current hardware limitations available. In fact, while the 3D kernels have a good ability to extract the spatial correlation of the MRI sequence, on the other they need a high capacity in GPU global memory size.
Thus, an excessive reduction of the MRI to adapt to the available hardware causes an inevitable
loss in segmentation details (i.e probable motivation of $92.5 \%$ dice). 
Therefore, we recommend future tests with 3D kernel at more higher MRI resolution with the help
of NVIDIA V100 GPUs \cite{ZheJia} that presents $16-32$ GB of global memory.

Furthermore, the concatenation of two CNNs, the second working at the full resolution cropping the region of interests containing the atria, is a good strategy that will minimize the impact of this choice. An alternative is to work with the information at two levels of detail, which has shown to achieve a better performance \cite{Zhaohan}.

Proposed V-FCNN simplifies the v-net \cite{Fausto} by removing the skip-paths, in an attempt to achieve a much quicker training convergence. The constitutive advantage of the skip-paths is to increase the localization, but this comes at the cost of a considerable slow down of the speed of the network (i.e GPU global memory leak) while propagating the gradient, at every iteration, forward and back through those paths \cite{AdamPaszke}. 

The choice of the loss function is an important consideration in any CNN, and the joint minimization of MSE and Dice Loss has been adopted in our solution. The interpretation is that the MSE looks at global volumetric features, while the Dice Loss (DL) regularize it trying to fit local details. This choice allowed the learning to avoid local minima and the corresponding slow convergence of using only MSE. The optimal weight between them, and the inclusion of further criteria such as an L1/L2 loss or a statistical distance to a library of existing cases, should be addressed in future extensions of this work. 

An alternative strategy to 3D kernels is the use of recurrent units \cite{Poudel},\cite{Jianxu},  where recurrence is used to capture the redundancy between adjacent slices. The use of recurrence is a much more memory efficient approach (i.e fewer parameters to capture recurring partners), also reporting improvements at the challenging apical slices of the left ventricle \cite{Poudel}. But these solutions are limited by the vanishing gradient that unfortunately occurs within long sequences, what can be partially avoided by imposing upper bound constraints on the backward gradient (ie. gradient clipping) or with a regularisation term \cite{Razvan}. Future work is thus still needed to maximize the synergies between 3D kernels and recurrent units within a more higher input MRI resolution.

\section*{Acknowledgements}

This work was supported by the Wellcome/EPSRC Centre for Medical Engineering at King’s College London [g.a. 203148/Z/16/Z]. PL holds a Wellcome Trust Senior Research Fellowship [g.a. 209450/Z/17/Z].
%------------------------------------------------------------------------
% Bibliography
%

%\bibliographystyle{unsrt}
%\bibliography{references} 

%\end{thebibliography}
\end{document}